\documentclass[pdflatex,sn-basic]{sn-jnl}

\usepackage{graphicx}%
\usepackage{multirow}%
\usepackage{amsmath,amssymb,amsfonts}%
\usepackage{amsthm}%
\usepackage{mathrsfs}%
\usepackage[title]{appendix}%
\usepackage{textcomp}%
\usepackage{manyfoot}%
\usepackage{booktabs}%
\usepackage{algorithmicx}%
\usepackage{algpseudocode}%
\usepackage{listings}%
\usepackage[table]{xcolor}%
\usepackage[linesnumbered,ruled,vlined]{algorithm2e}%
\usepackage[export]{adjustbox}%

\usepackage{graphicx}
\usepackage{tikz}
\usepackage{multicol}

\usepackage{array} 
\usepackage{color}
\usepackage{colortbl}
\usepackage{xcolor}

\newcommand{\applyColor}[1]{
  \ifnum#1>75
    \cellcolor{red!50}
  \else
    \ifnum#1>50
      \cellcolor{yellow!50}
    \else
      \ifnum#1>25
        \cellcolor{green!50}
      \else
        \cellcolor{white}
      \fi
    \fi
  \fi
  #1
}

\definecolor{customblue}{RGB}{165, 195, 255}

\newcommand{\customcell}[2]{\cellcolor{customblue!#1}{#2}}

\begin{document}

\title[Article Title]{A Unified Manifold Similarity Measure Enhancing Few-Shot, Transfer, and Reinforcement Learning in Manifold-Distributed Datasets}


\author[1]{\fnm{Sayed Waleed} \sur{Qayyumi}}\email{s.qayyumi@westernsydney.edu.au}

\author[2]{\fnm{Laurence A. F} \sur{Park}}\email{l.park@westernsydney.edu.au}

\author[3]{\fnm{Oliver} \sur{Obst}}\email{o.obst@westernsydney.edu.au}

\affil*[1]{\orgdiv{Centre for Research in Mathematics and Data Science,
School of Computer, Data and Mathematical Sciences}, \orgname{Western Sydney University}, \orgaddress{\street{Locked Bag 1797, Penrith}, \city{Sydney}, \postcode{2751}, \state{NSW}, \country{Australia}}}


\abstract{Training a classifier with high mean accuracy from a manifold-distributed dataset can be challenging. This problem is compounded further when there are only few labels available for training. For transfer learning to work, both the source and target datasets must have a similar manifold structure. As part of this study, we present a novel method for determining the similarity between two manifold structures. This method can be used to determine whether the target and source datasets have a similar manifold structure suitable for transfer learning. 

We then present a few-shot learning method to classify manifold-distributed datasets with limited labels using transfer learning. Based on the base and target datasets, a similarity comparison is made to determine if the two datasets are suitable for transfer learning. A manifold structure and label distribution are learned from the base and target datasets. When the structures are similar, the manifold structure and its relevant label information from the richly labeled source dataset is transferred to target dataset. We use the transferred information, together with the labels and unlabeled data from the target dataset, to develop a few-shot classifier that produces high mean classification accuracy on manifold-distributed datasets.

In the final part of this article, we discuss the application of our manifold structure similarity measure to reinforcement learning and image recognition.}




\maketitle

\section{Introduction}
\label{sec1}
A few-shot learning method is used in machine learning and artificial intelligence to train models with a limited number of examples. In contrast to traditional machine learning, few-shot learning utilises prior knowledge from related tasks in order to gain insight from scarcely labeled data. Traditional machine learning requires a large amount of labeled data in order to be accurate; however, in real-world scenarios, it is often impossible or prohibitively expensive to acquire extensive labeled datasets for each new task or category. The goal of few-shot learning is to develop algorithms and techniques that allow models to learn from a limited number of examples per class in an efficient manner. In contrast to traditional machine learning, few-shot learning enables machines to learn as humans do, by generalising information from past experiences to new situations even when only a few examples are provided. It is therefore fair to say that few-shot learning bridges the gap between traditional machine learning and human-like learning abilities. 

The process of transfer learning in machine learning and deep learning refers to the use of knowledge gained from one task to improve performance on another. Instead of starting from scratch, transfer learning uses pre-trained models or representations to accelerate training, enhance generalisation, and improve performance on target tasks. Traditionally, machine learning models are trained on specific datasets for specific tasks, which requires a considerable amount of labeled data. Annotating large datasets can, however, be time-consuming, expensive, or impractical in certain circumstances. A transfer learning method can overcome this limitation by transferring knowledge from an abundantly labeled source domain to a scarcely labeled target domain. Transfer learning is based on the principle that knowledge gained from solving one task can be applied to solving another related task. Utilising this knowledge, models can extract meaningful features, capture important patterns, and generalise well to new data even when limited labeled data is available.

Manifold distributed datasets are those whose samples are located on or near a low-dimensional manifold within a high-dimensional space. In traditional datasets, data points are often assumed to be independent and identically distributed (i.i.d.). However, real-life data often demonstrate complex structures and correlations that can be better understood by considering the underlying manifold. A manifold is a curved or folded surface embedded in a higher-dimensional space. The manifold represents the underlying structure of data, where each point represents a separate sample. By mapping high-dimensional data points to a lower-dimensional manifold representation, manifold learning uncovers the intrinsic structure and geometry of the data.

A manifold-distributed dataset can be expressed mathematically as follows:
Let $\mathcal{M}$ be a smooth $d$-dimensional manifold embedded in a higher-dimensional space $\mathbb{R}^D$, where $d < D$. The manifold-distributed dataset $\mathcal{D}$ consists of $N$ data points, each denoted as $x_i \in \mathbb{R}^D$, where $i = 1, 2, ..., N$. Here's a detailed explanation:
\begin{itemize}
    \item Manifolds represent curved surfaces embedded in a higher-dimensional space that are similar to Euclidean spaces at a local level, but are complex at a global level.
    
    \item In the context of datasets, a manifold-distributed dataset refers to a collection of data points that are distributed approximately on or close to a lower-dimensional manifold within a higher-dimensional ambient space.
    
    \item $\mathcal{M}$ is assumed to be smooth, which implies that it does not have sharp edges or singularities. Furthermore, it is $d$-dimensional, meaning that $d$ independent variables can be used to parameterise it locally.

    \item In the dataset $\mathcal{D}$, each data point $x_i$ is represented as a vector $\mathbb{R}^D$, where $D$ represents the dimension of the ambient space, which is typically much greater than the intrinsic dimension of the manifold $d$.

    \item The dataset $\mathcal{D}$ is considered to be manifold-distributed because the data points are not randomly distributed in the ambient space, but instead tend to cluster around the smaller manifold $\mathcal{M}$.

    \item Manifold-distributed datasets pose a challenge because they require uncovering the underlying structure of the manifold $\mathcal{M}$, often using techniques like dimensionality reduction, manifold learning, and topology inference.
    
\end{itemize}
In summary, a manifold-distributed dataset consists of a collection of data points that are approximately arranged on or near a lower-dimensional manifold within a higher-dimensional ambient space, presenting both challenges and opportunities for analysis and interpretation.

Reinforcement learning and few-shot learning are two distinct machine learning paradigms that each come with their own challenges and techniques. Nevertheless, they can be effectively combined, especially when combined with transfer learning principles, for the purpose of handling complex learning tasks involving limited labeled data, especially when combined with transfer learning principles.

During reinforcement learning, agents interact with their surroundings, take actions, and then receive feedback from the environments as they learn a policy over time. The goal of reinforcement learning is to maximise cumulative rewards. Q-learning, deep Q-networks, policy gradient methods, and other RL algorithms are all used to teach agents how to make optimal decisions in dynamic and uncertain environments.

Reinforcement learning and few-shot learning can be bridged using transfer learning. When an agent is pre-trained in a diverse set of environments or tasks, it can acquire generalisable skills and policies that can be adapted to new environments with a relatively small amount of training data. Researchers and practitioners can develop robust and adaptive learning systems by integrating transfer learning principles into reinforcement learning and few-shot learning frameworks that can learn effectively from small amounts of data and generalise well to new and unknown situations by integrating transfer learning principles. In real-world situations where data scarcity and adaptability are key challenges, this synergy between reinforcement learning, few-shot learning, and transfer learning opens up new possibilities.

A novel method is presented in this paper for determining the similarity between two manifold structures. To determine whether target and base datasets have similar manifolds and are suitable for transfer learning, this method can be used. Using the mentioned method, the paper then presents a novel algorithm for few-shot learning of manifold distributed datasets. Using a similar source labeled manifold distributed dataset, the algorithm learns the manifold structure of the data. Then, using the manifold structure learned from the source dataset, along with the labeled and unlabeled data from the target manifold distributed dataset, a few-shot learning classifier is trained. Finally, we describe how similarity in manifold structures can be used for reinforcement learning.

This article makes the following contributions:
\begin{itemize}
\item A novel method for calculating the similarity between two manifold structures is proposed. The purpose of this method is to determine whether a base dataset is suitable for transfer learning. (Section~\ref{simi}) 
\item A novel algorithm is presented for the transfer learning when data is manifold distributed and labels are limited in few-shot learning scenarios. Transfer learning is achieved using graphs and random walks. (Section~\ref{algo}).
\item Lastly, we discuss the application of our approach to the similarity of manifold structures and few-shot learning for reinforcement learning. (Section~\ref{RL}).  

\end{itemize}
This article will proceed in the following manner:
Section~\ref{sec:related} examines current research on few-shot learning, transfer learning, and classification of manifold-distributed data with few labeled samples. Section~\ref{simi} presents a method to compare the similarity of manifold structures between two manifold distributed datasets. Section~\ref{algo} outlines our proposed algorithm. Section~\ref{exper} describes the experiments conducted on synthetic and real-life datasets. Our results are also compared with other state-of-the-art methods in this section. Section~\ref{RL} discusses the application of similarity and few-shot learning to reinforcement learning using our proposed method.

\section{Background and related work}
\label{sec:related}
The semi-supervised learning paradigm is ideal when a large amount of unlabeled data is available alongside a small quantity of labeled data~\cite{reddy2018semi,zhu2005semi}. The traditional method of supervised learning involves training models solely on labeled data, which is time-consuming and costly. On the other hand, semi-supervised learning emphasises the inclusion of additional information from unlabeled data to enhance the generalisation and performance of the model. 

Using the vast amount of unlabeled data that is available is a major challenge for semi-supervised learning. Various techniques are employed to overcome this obstacle. You can use self-training~\cite{zoph2020rethinking}, a process in which unlabeled data points are labeled iteratively and added to the training set, or co-training~\cite{zhang2014semi}, where multiple views of data are used to improve learning. The accuracy and robustness of semi-supervised learning algorithms are improved by learning from both labeled and unlabeled data.

Recent advances in deep learning, meta-learning, and transfer learning have led to significant progress in few-shot learning. In order to address the few-shot learning problem, numerous approaches and techniques have been developed, each with its own strengths and characteristics. Machine learning approaches such as few-shot learning~\cite{wang2020generalizing} address the problem of training models with a limited number of labeled examples (shots) per class or task. In real-world applications, such scenarios are common where it is difficult or expensive to collect large, labeled datasets. Few-shot learning allows models to generalise well to new, unknown classes or tasks with very few examples.

The challenges associated with Few-shot learning have been addressed through the development of a number of strategies and techniques. Few-shot learning methods that utilise meta-learning~\cite{elsken2020meta} are trained on a variety of tasks, each accompanied by a limited number of labeled examples. The models are taught how to develop an effective initialisation or parameter updating rule. This is to facilitate its ability to adapt rapidly during testing and inference to new tasks or classes.

A common strategy used in few-shot learning is meta-learning~\cite{jiang2020multi}, also known as learning to learn. Through the use of meta-learning algorithms, models can rapidly adapt to new tasks with limited labeled data by accumulating knowledge from multiple related tasks during a pre-training phase. Models become more adept at generalising to new classes as they learn how to learn. Metric learning attempts to learn a similarity metric that measures the similarity or dissimilarity between data points in order to make accurate predictions based on limited data. 

Transfer learning~\cite{rostami2019deep,qayyumi2023few} is an approach to improve performance on new, related tasks using the knowledge gained from pre-trained models on large datasets (pre-training). In this article we present a few-show classifier using transfer learning. 

Memory-augmented architectures have also been shown to be promising for few-shot learning. Both training and testing are conducted using models that use external memory structures as a dynamic storage medium. Learning through few-shots is made possible by the ability to retain important knowledge and adapt it to new situations. By utilising data augmentation techniques~\cite{zhou2021flipda}, data is augmented artificially by rotating, translating, scaling, and introducing noise. 
Further, few-shot learning has been applied to improve model performance, adaptability, and generalisation through the use of generative modeling, attention mechanisms, and knowledge distillation. A variety of generative models, including Generative Adversarial Networks (GANs)~\cite{robb2020few} and Variational Autoencoders (VAEs), can be used to generate new data samples, thereby improving the available labeled data for training. 
In addition to improving performance and reducing over-fitting, ensembling methods~\cite{dvornik2019diversity} can also enhance performance.

Metric learning~\cite{jiang2020multi} is another popular approach to few-shot learning. A metric-based method seeks to infer a similarity metric or embedding space in which samples from the same class are more similar than samples from different classes. Using these methods, effective classification and recognition can be achieved with a minimum amount of training data.

In recent years, several influential papers have significantly advanced the few-shot learning technique. Several innovative algorithms and methodologies are presented that have pushed the boundaries of what is possible even with limited training examples. Here are some interesting ideas presented in a number of popular papers. In~\cite{vinyals2016matching} the authors introduced matching networks for one-shot learning. The authors propose a trainable model that compares examples from a support set with examples from a target set. Similarities between samples are computed using a differentiable nearest-neighbor algorithm. As demonstrated in this study, memory-augmented architectures can be useful for few-shot learning.

In~\cite{snell2017prototypical} the authors presented a few-shot learning approach utilising prototypical networks. The algorithm proposed is based on the learning of a metric space where samples from the same class are closer together than samples from other classes. By representing classes with learned prototypes, this method enables efficient inference and generalisation of new classes.~\cite{ren2018meta} introduced the concept of meta-learning using few-shot learning as a context. In order to facilitate rapid adaptation to new tasks with limited labeled data, the authors propose a meta-learning algorithm that learns the initialisation of the model's parameters as the task progresses. As a result of training on multiple related tasks, generalisation and adaptation to new classes are enhanced.

In~\cite{finn2017model} the authors introduced the model-agnostic meta-learning (MAML) algorithm, which provides a general framework for few-shot learning. The use of MAML allows models to be quickly adapted to new tasks with limited data by optimising their parameters. MAML learns from few examples efficiently by iteratively updating its initialisation based on task-specific gradients.

The transfer learning technique is a powerful tool in machine learning and computer vision for improving the performance of related tasks based on the knowledge gained from the first task. The transfer of representations, features, or knowledge from one domain to another reduces the requirement for expensive training processes and large labeled datasets. Models are believed to be capable of transferring knowledge gained from one task to another, improving performance, generalisation, and convergence. 

A number of influential papers have contributed to the development and advancement of transfer learning techniques. In these papers, several approaches and architectures have been proposed which have had a significant impact on the field. In~\cite{krizhevsky2012imagenet} the authors used deep convolutional neural networks (CNNs) to introduce the concept of transfer learning. The study demonstrated the effectiveness of pre-training CNNs on a large-scale dataset (ImageNet) and fine-tuning them for specific target tasks. A large-scale dataset could be generalised to different tasks, resulting in significant improvements in accuracy.

In~\cite{simonyan2014very} the authors introduced VGGNet, which achieved excellent performance on the ImageNet challenge. The study demonstrated that transfer learning is effective when it is applied to very deep networks. According to the authors, VGGNet learns rich representations on ImageNet that are generalisable to other tasks after pre-training.

In~\cite{he2016deep} the authors introduced the ResNet architecture, which addresses the problem of training deep neural networks by introducing residual connections. Using ResNet, the authors demonstrated superior performance on ImageNet and highlighted the importance of deep networks in transfer learning. The network could learn residual mappings by using residual connections, allowing features from pre-trained models to be reused more effectively. 

Goodfellow et al.~\cite{goodfellow2020generative}, despite not exclusively addressing transfer learning, introduced the concept of generative adversarial networks (GANs), which are widely used for transfer learning. An adversarial GAN consists of a generator and a discriminator network. Generators learn to generate synthetic samples that cannot be distinguished from real samples, while discriminators learn to distinguish real samples from fake ones. GANs have been used for a variety of transfer learning tasks, including domain adaptation and style transfer. 

In few-shot learning using manifold-distributed data, it is difficult to determine the manifold structure of the data accurately due to the limited number of observations available for training. There is no mathematical possibility of estimating the manifold structure in zero-shot and one-shot learning, for instance. Consequently, it is critical to determine the minimum number of samples required to accurately determine the manifold structure. It may not be possible to capture the nonlinear, complex relationships between manifold distributed data using a small number of examples. Consequently, the model may be overfit or underfit, resulting in poor performance when new, unseen data is introduced. Transfer learning success is heavily dependent on selecting a suitable source dataset. However, the algorithm used also contributes to this success~\cite{ganin2016domain}.

Another challenge is that few-shot learning assumes that the training examples are representative of the entire manifold distribution, which is not necessarily true for manifold distributed data. When the training data is biased towards particular parts of the manifold or does not cover the full range of variations in the data, a model may fail to generalise to new data points that are not included in the training set. 

Furthermore, when using few-shot learning with manifold distributed data, it is essential to select the appropriate distance metric or similarity metric. The selection of an appropriate distance metric can be challenging due to the fact that different manifolds require different distance metrics.

Finally, few-shot learning may require complex and computationally expensive models if there is a large amount of distributed data. The training data for these models must be collected in large quantities in order to avoid overfitting, which is in conflict with the few-shot approach.

Manifold-distributed data can be represented intuitively and naturally with graphs. Edges represent pairwise relationships between data points, while nodes represent data points. Graphs help capture the complex, nonlinear relationships between data points of manifold distributed data. Graphs can also be used to incorporate additional information about the data into the classification process, such as pairwise similarities, distances, or labels. Pairwise relationships between data points can be used to assign edge weights. In summary, graphs offer an effective framework for addressing many of the challenges posed by manifold distributed data classification~\cite{cai2010graph}. Using the constructed graph, we can determine the likelihood that unlabeled observations belong to a particular class based on their proximity to labelled observations. Graph-based classification has proven to be a state-of-the-art approach to classifying manifold distributed data~\cite{tu2016graph,tu2014novel}.

\section{Measuring similarity of manifolds}
\label{simi}
There are a number of general issues and challenges associated with measuring similarity in various fields, such as machine learning, data mining, information retrieval, and natural language processing. One of the primary challenges in measuring similarity is choosing the appropriate similarity metric and distance measure. Different metrics can result in inconsistent similarity assessments. When comparing text data, for example, Euclidean distance and cosine similarity can produce different similarity scores.

Dimensionality is another issue. As the number of dimensions increases, the amount of data needed to maintain meaningful similarity between points increases exponentially. As a result, it can be difficult to determine the true similarity between high-dimensional data points accurately. Also, handling noisy or missing data can impact similarity measurements.

Moreover, similarity can be both subjective and context-dependent, with some similarities occurring in one context and not in another, thus emphasising the importance of domain-specific understanding and feature selection when attempting to measure similarity.

Manifolds, which are geometric structures that are locally similar to Euclidean space, but may have complex global shapes, present additional challenges when measuring similarity. Due to their inherent geometric properties, traditional similarity measures may not be suitable for manifolds.

For addressing these challenges, researchers have developed techniques such as manifold learning algorithms, which aim to uncover the underlying structure of the data, reduce its dimensions, and retain its essential characteristics while reducing its dimensionality. Isomap, Locally Linear Embedding (LLE), and t-distributed Stochastic Neighbor Embedding (t-SNE) are among the algorithms available for measuring similarity in manifold spaces.

There are, however, some limitations to these solutions. There is often a need to tune parameters in manifold learning algorithms, and the choice of parameters can significantly alter the results. Furthermore, these methods are not always effective with high-dimensional data or datasets with sparse or noisy samples. As the geometric relationships in manifold spaces are not always in alignment with intuitive notions of similarity, it can be challenging to interpret similarity measurements.

Measurement of the similarity between two datasets and manifolds is crucial to determining the efficiency and feasibility of knowledge transfer. Using knowledge from one domain, typically called the source domain, transfer learning is aimed at improving the performance of a model in another, but related, domain, called the target domain. For a deeper understanding of the underlying structures and distributions in both domains, it is necessary to assess the similarity between datasets. Transfer learning is more effective when datasets have similar statistical characteristics, such as the distribution of data, the representation of features, and the semantic relationships among them. It is possible to effectively transfer knowledge from the source domain to the target domain, enhancing generalization and performance.

It is also crucial to measure the similarity between manifolds, which represent the intrinsic low-dimensional structures of data. In high-dimensional data, manifolds can be used to capture geometric patterns and relationships, which are necessary for understanding the underlying structures of target domains and source domains. By analyzing the geometric similarities between manifolds, we can determine whether data in both domains have similar geometric structures, allowing for the transfer of meaningful and relevant knowledge.

Several works have stressed the importance of assessing datasets and the many similarities between them when it comes to transfer learning. In~\cite{cao2010adaptive}, the authors give a comprehensive overview of transfer learning techniques and emphasize the importance of selecting the appropriate source and target domains on the basis of similarity measures. In addition,~\cite{weiss2016survey} present a variety of transfer learning paradigms and emphasize the importance of measuring dataset similarity in order to achieve successful knowledge transfer.

In order to achieve effective transfer learning, it is essential to measure the similarity between datasets and manifolds. Understanding the statistical and geometric relationships between domains can help us improve model performance in the target domain. Taking into account the above, we examined various similarity measures in order to select datasets from the source domain that would be most suitable for transfer learning.  Various measures can be used to measure the similarity between datasets from a source and a target domain. However, these measures do not work when data is manifold distributed. Therefore, we examined a measure that can be effective when data from the source domain and the target domain are manifold distributed. 

\begin{figure}[htb]
    \centering
    \includegraphics[width=1\textwidth]{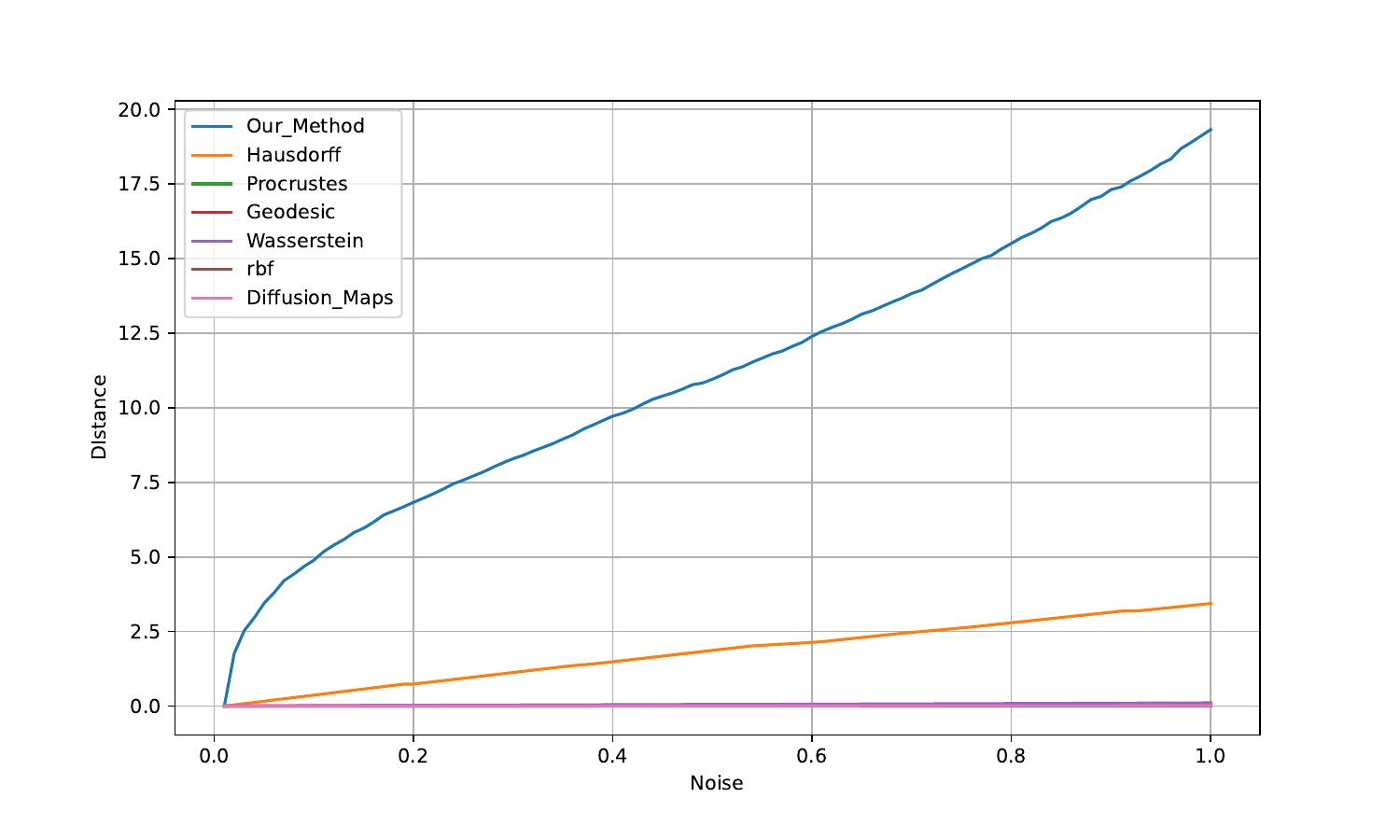}
    \caption{Distance calculated using various methods as more noise is added to one of the manifolds}
    \label{fig:distancemeasures}
\end{figure}

A graph is a natural mechanism for unraveling the structure of a manifold~\cite{tu2016graph,vishwanathan2010graph}. Therefore, we examined a measure based on graphs. Distance can be measured by random walks over graphs, and that is what we are doing in our distance measurement.  We have presented our approach to measuring distance between two manifold distributed datasets in 
Algorithm~\ref{alg:random_walk_distance}. To determine the effectiveness of our measure, we compared it to all potential measures that can be used in transfer learning. We used both synthetic and real-world datasets. A comparison of our method of measuring similarity with all other methods using synthetic data of a Swiss roll is presented in Figure~\ref{fig:distancemeasures}. We began by comparing identical Swiss rolls. We continued to measure similarity by keeping one Swiss roll the same and adding normal distributed noise to the other Swiss roll. Figure~\ref{fig:distancemeasures} illustrates the results of the experiments.

\begin{figure}[htb]
    \centering
    \includegraphics[width=12cm, height = 10cm]{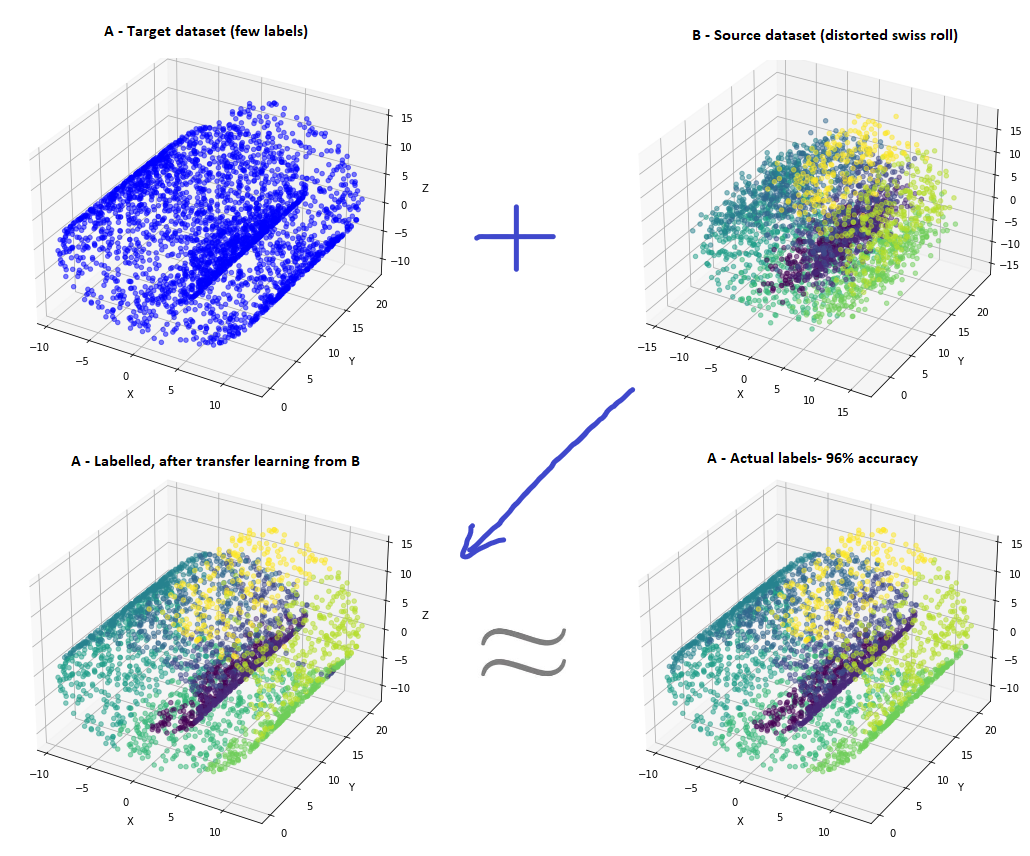}
    \caption{Transfer Learning - Swiss roll manifold}
    \label{fig:transferlearning}
\end{figure}

We also evaluated the performance of our approach to similarity using images from the miniimagenet dataset. In Table~\ref{tab:avgdis}, a single image of each species is compared with 50 images of other species and its own. Using the set of 50 images, similarity is determined by their average distance (using our method). Based on our results, we are able to identify similar manifold structures and distinguish between different manifolds based on similarity of their structures. 

In Table~\ref{tab:dismeasure}, we compare our method with other similarity measures. In the results, we can see that similarity decreases as we compare an image of a bird with other bird-like species and with non-bird-like species. In contrast, other measures of similarity are not able to accomplish this.  When we compare similar manifolds between two objects and the number of manifolds is the same, our approach to comparing similarity between images works well. Our method, for example, works when comparing a duck image to another duck image, but fails when comparing a duck image with an image containing more than one duck. The reason for this is that the manifold structure of an image of a single duck differs from the manifold structure of an image with many ducks.

\begin{table}[htb]
  \centering
  \setlength{\abovecaptionskip}{12pt}
\caption{MiniImageNet Dataset- A single image of each species is compared with 50 images of other species and its  own. Similarity is based on the average distance (using our method) from the set of 50 images per class.}
    \label{tab:avgdis}
  \begin{tabular}{l*{6}{c}}
    \hline
    & Birds(50) & Ducks(50) & Toco(50) & Snakes(50) & Dogs(50) & Lions(50) \\
    \hline
    \includegraphics[width=1.5cm, height=1cm]{bird_Main.jpeg} & \customcell{100}{384.42} & \customcell{75}{411.45} & \customcell{70}{531.44} & \customcell{30}{1474.24} & \customcell{10}{1967.65} & \customcell{50}{956.75} \\
    \includegraphics[width=1.5cm, height=1cm]{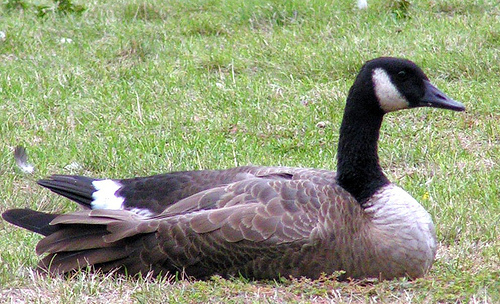} & \customcell{50}{627.5} & \customcell{100}{187.62} & \customcell{60}{477.95} & \customcell{20}{1597.2} & \customcell{30}{1222.55} & \customcell{10}{1739.42} \\
    \includegraphics[width=1.5cm, height=1cm]{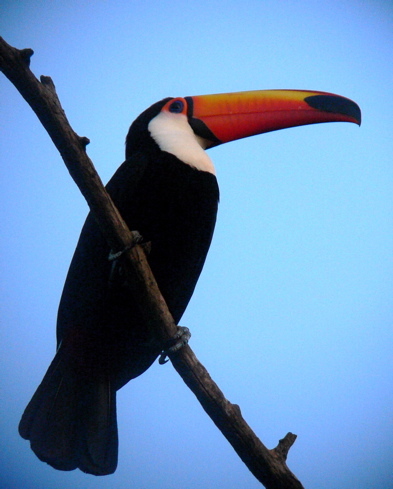} & \customcell{75}{576.21} & \customcell{60}{596.57} & \customcell{100}{385.16} & \customcell{30}{801.92} & \customcell{10}{863.38} & \customcell{40}{752.33} \\
    \includegraphics[width=1.5cm, height=1cm]{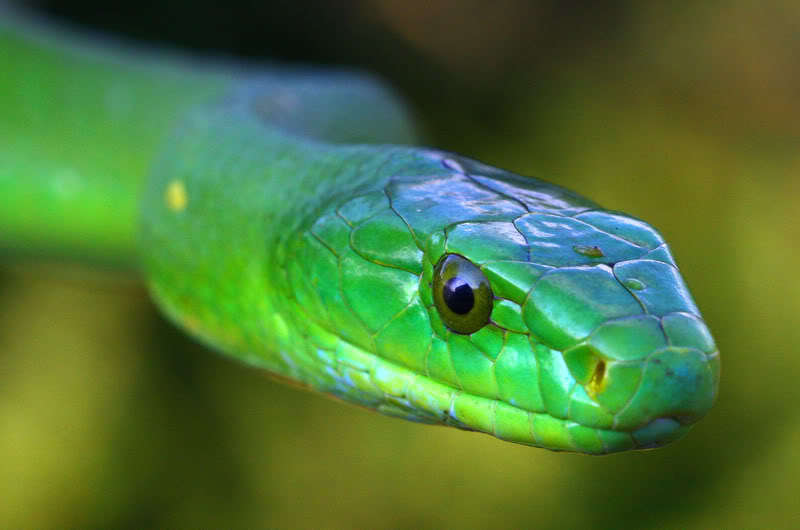} & \customcell{60}{1392.87} & \customcell{45}{1455.15} & \customcell{10}{1660.04} & \customcell{100}{609.33} & \customcell{30}{1461.78} & \customcell{10}{1683.36} \\
    \includegraphics[width=1.5cm, height=1cm]{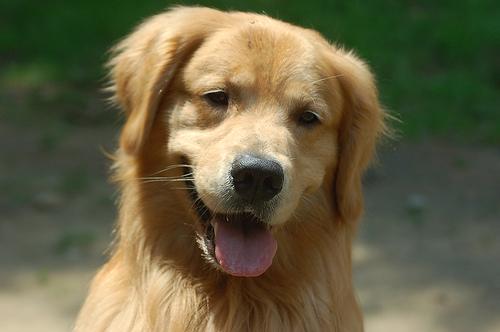} & \customcell{40}{1323.62} & \customcell{50}{870.06} & \customcell{10}{2908.67} & \customcell{30}{1674.11} & \customcell{100}{332.47} & \customcell{60}{669.92} \\
    \includegraphics[width=1.5cm, height=1cm]{lion.jpeg} & \customcell{20}{2156.54} & \customcell{10}{2423.52} & \customcell{20}{2189.73} & \customcell{10}{2569.52} & \customcell{60}{384.42} & \customcell{100}{310.96} \\
    \hline
  \end{tabular}
\end{table}

Graph similarity is measured by comparing the random walk matrices of two graphs over time $t$. Using the adjacency matrices of both graphs and their respective node counts, the function constructs the random walk matrices. The measure of similarity is calculated by computing the Frobenius norm of the difference between the random walk matrices of the two graphs. Inherently, this method calculates the similarity between adjacent matrices by scanning the rows.

It is possible, however, to scan the columns instead of the rows by transposing the adjacency matrices before performing similarity calculations. By transposing the matrices, we effectively switch from a row-based perspective to a column-based perspective, allowing us to scan columns while retaining the fundamental nature of the random walk distance calculation.

Additionally, we can combine the row and column perspectives into one matrix to measure similarity using a concatenation of the rows and columns. 

We conducted experiments on all of the above-mentioned methods. The results of these experiments are shown in Table~\ref{tab:diffmethods}, which provides insight into how effective and accurate each method is in measuring similarity. For the transfer learning approach, we chose row scanning because it is more efficient.

\begin{table}[htb]
  \centering
  \setlength{\abovecaptionskip}{12pt}
\caption{MiniImageNet Dataset- We compare a single image of each species with ten images of its own species. Similarity is determined by using the average distance between the main image and a set of 10 images of the same species. This is done using the rows, columns, and row and column methods. A higher blue density indicates less similarity.}
    \label{tab:diffmethods}
  \begin{tabular}{l*{3}{c}}
    \hline
    & Based on Rows & Based on Columns & Based on Column and Rows  \\
    \hline
    \includegraphics[width=1.5cm, height=1cm]{bird_Main.jpeg} & \customcell{40}{79.99} & \customcell{30}{77.35} & \customcell{100}{150.22}  \\
    
    \includegraphics[width=1.5cm, height=1cm]{notbird2.jpeg} & \customcell{20}{17.97} & \customcell{50}{71.19} & \customcell{100}{331.40}  \\
    \includegraphics[width=1.5cm, height=1cm]{toco.jpeg} & \customcell{30}{69.00} & \customcell{40}{80.73} & \customcell{80}{92.34}  \\
    
    \includegraphics[width=1.5cm, height=1cm]{dog.jpeg} & \customcell{20}{24.47} & \customcell{70}{92.07} & \customcell{100}{100.98}  \\
    \includegraphics[width=1.5cm, height=1cm]{lion.jpeg} & \customcell{30}{29.50} & \customcell{30}{33.61} & \customcell{80}{96.08}  \\
    \hline
  \end{tabular}
\end{table}

\section{Transfer Learning}
\label{algo}
We describe our approach to transfer learning between two manifold distributed datasets $X1$ and $X2$ with labels $y1$ and $y2$ in algorithm~\ref{alg:swiss_roll_classification}. Target dataset $X1$ is a manifold distributed dataset with a few labels $y1$ whereas source dataset $X2$ is another manifold distributed dataset with many labels $y2$. The classifier is trained using the source dataset and a portion of the target dataset that is labeled. There are three parts to the entire method. 

Algorithm~\ref{alg:random_walk_distance} calculates distances between two manifold structures. The method serves as a distance measure for the classifier we use, which is the $k$ neighbor classifier. Furthermore, the method determines whether the source dataset has a similar manifold structure to the target dataset and can be used for transfer learning. For transfer learning to be effective, it is recommended that the source dataset has a level of manifold similarity. When we calculate similarity, we use the row of the metrics, not the column. Using rows or columns will not change the results as such, but the calculation may differ to some extent. 

Algorithm~\ref{alg:construct_graph} constructs graphs of manifold distributed data. A random walk distance measure is used in training classifiers using these graphs. A random walker walks over these graphs, distance is calculated which is used by the $k$ neighbor classifier to determine which neighbor is closest. 

Algorithm~\ref{alg:swiss_roll_classification} performs transfer learning. Using algorithm~\ref{alg:random_walk_distance} and ~\ref{alg:construct_graph}, this algorithm trains a $k$ neighbor classifier using random walk as the distance metric. In this algorithm, the knowledge of manifold structure (source dataset) and the relationship between manifolds and labels (source dataset) is transferred to the target dataset. In conjunction with the limited labels available in the target dataset, these information are used to train a classifier that is then used to classify the unlabeled data in the target dataset.

\begin{algorithm}[htb]
\caption{Measuring Similarity}
\label{alg:random_walk_distance}
\KwData{adj1, adj2, t}
\SetKwInOut{Input}{Input}
\SetKwInOut{Output}{Output}
\Input{adj1, adj2, t}
\Output{distance}
\BlankLine
$\text{Get and assign to $n1,n2$ number of rows in } \textit{adj1,adj2}$\;

\text{Create identity matrices of size}  $n1 \times n1$ \text{and} $n2 \times n2$ \;

\text{Create weight matrices W1, W2} $\colon W1 =$  \text{Inverse of } $(I1 - t \times \textit{adj1})$  \text{and} $W2 = $ \text{Inverse of } $(I2 - t \times \textit{adj2})$ \;

\text{Calculate distance} $d = \text{norm of } (W1 - W2)$\;

\Return $d$\;
\end{algorithm}

\setcounter{AlgoLine}{0}
\begin{algorithm}[htb]
\caption{Construction of Graphs}
\label{alg:construct_graph}
\KwData{dataset d, k\_neighbors}
\SetKwInOut{Input}{Input}
\SetKwInOut{Output}{Output}
\Input{dataset d, k\_neighbors}
\Output{graph g}
\BlankLine
\text{Create a new empty graph}\;

Calculate the Euclidean distance matrix for $d$\;

\For{\textit{i} \textbf{in} \textbf{range}(\textit{length}(\textit{d}))}{
    \textit{neighbors} $\leftarrow$ Indices of \textit{k\_neighbors} nearest neighbors of data point \textit{i}\;
   
    \For{\textit{j} \textbf{in} \textit{neighbors}}{
        Add an edge between data point \textit{i} and data point \textit{j} in \textit{g}\;
    }
}

\Return \textit{g}\;
\end{algorithm}

\setcounter{AlgoLine}{0}
\begin{algorithm}[htb]
\caption{Transfer Learning}\label{alg:swiss_roll_classification}
\KwData{X1, y1, X2, y2, k,DT}
\SetKwInOut{Input}{Input}
\SetKwInOut{Output}{Output}

\Input{X1, y1, X2, y2, k,DT}
\Output{$\mu\alpha $ (mean accuracy of classification)}

Assign random walk distance between $X1,X2$ to $d$\;

\If{$d$ $\leq$ $DT$ }{

Scale the features of $X1,X2$ to the range [0, 1]\;

Construct graphs $g1,g2$ from scaled $X1,X2$ with \textit{k} nearest neighbors\;

Convert $g1,g2$ to adjacency matrices\;

Create a new $k$ Neighbors classifier with $k$ neighbors and Algorithm~\ref{alg:random_walk_distance} as distance metric\;

Train classifier on $X2$ and $X1\_labeled$ with labels $y2,y1$\;

Classify $X1\_unlabeled$ using the trained classifier\;

\Return Prediction ($y1$)\;
}
\end{algorithm}

We have provided a detailed explanation of the transfer learning process in~\ref{fig:transferlearning}. There are two Swiss roll manifolds $A$ and $B$. $A$ is a Swiss roll dataset with very few labels (color class). $B$ is a distorted Swiss roll dataset, but all labels are available. There is a similarity in the manifold structure of both datasets. Using the~\ref{alg:random_walk_distance} we are able to measure the similarity between these datasets and determine if we should use them as a learning resource. Considering that the manifold in this case is very similar ($70\%$), we have employed it for transfer learning. It is evident from the output (A-labelled) that transfer learning has added a great deal of value in this instance, achieving an accuracy of $96\%$ when compared to the actual labels. 

\section{Experimental Results}
\label{exper}
We conducted experiments on both synthetic and real-world manifold distributed datasets. These experiments were conducted to evaluate the effectiveness of our method and determine when we should stop using transfer learning in few-shot learning with manifold distributed datasets. Few-shot learning with manifold distributed datasets makes it extremely difficult to construct a classifier with a high mean accuracy. Therefore, it is logical to learn from a source dataset and transfer knowledge to a target dataset. 

There is a breakeven point after which transfer learning does not add value, and may even not be necessary. There may or may not be a need for transfer learning depending on the number of labels per class available in the target dataset and how different the source dataset is in terms of its manifold structure. In order to simulate this scenario, we experimented with $10, 20, 30$ and $40$ labels per class for synthetic and real-world datasets. We add noise to the source dataset as we increase the number of labelled data points per class. We begin with a noise level of 1 and increase the noise level by 1 every time. It is worth noting that an increase in noise represents an increase in standard deviation. As noise increases, standard deviation increases as follows: $0 \rightarrow 1 = 0.078$, $0 \rightarrow 2 = 0.29$, $0 \rightarrow 3 = 0.64$, and $0 \rightarrow 4 = 1$. A target dataset is created by removing all labels except the number required per class. An identical copy of the same dataset is used as the base dataset. Noise is added to all features except the class.

\begin{table}[htb]
    \centering
        \setlength{\abovecaptionskip}{12pt}
    \caption{MiniImageNet Dataset: The image of a bird is compared to that of other birds and other species based on our approach to similarity and other distance measures }
    \label{tab:dismeasure}
    \begin{tabular}{cc*{5}{>{\raggedleft\arraybackslash}p{1cm}}c}
        \toprule
        \textbf{A} & \textbf{B} & \textbf{ours} & \textbf{cos} & \textbf{rbf} & \textbf{pd} & \textbf{wd} & \textbf{hausdorff} \\
        \midrule
        \includegraphics[width=1.5cm, height=1cm]{bird_Main.jpeg} & \includegraphics[width=1.5cm, height=1cm]{bird_Main.jpeg} & 0.00 & 0.00 & 20.00 & 0.00 & 0.00 & 0 \\
        \includegraphics[width=1.5cm, height=1cm]{bird_Main.jpeg} & \includegraphics[width=1.5cm, height=1cm]{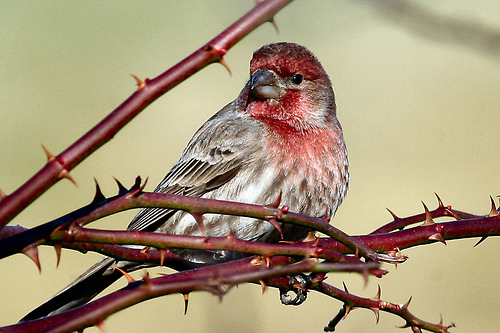} & 253.61 & 0.10 & 0.00 &  0.45 & 77.37 & 57066 \\
        \includegraphics[width=1.5cm, height=1cm]{bird_Main.jpeg} & \includegraphics[width=1.5cm, height=1cm]{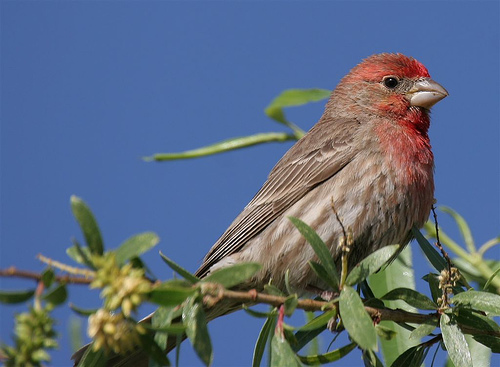} & 582.62 & 0.15 & 0.00 &  0.42 & 31.00 & 77164 \\
        \includegraphics[width=1.5cm, height=1cm]{bird_Main.jpeg} & \includegraphics[width=1.5cm, height=1cm]{notbird2.jpeg} &  2187.30 & 0.11 & 0.00 & 0.45 & 73.44 & 72403 \\
        \includegraphics[width=1.5cm, height=1cm]{bird_Main.jpeg} & \includegraphics[width=1.5cm, height=1cm]{notbird1.jpeg} &  4491.55 & 0.15 & 0.00 & 0.61 & 23.94 & 63515 \\
        \bottomrule
    \end{tabular}

\end{table}

\subsection{Synthetic datasets}
Several experiments were conducted using the data generated for Swiss roll, moon shape, and S curve shape manifolds. We generated two identical datasets: one as a target and the other as a source. The target dataset is modified by removing all labels except the number of labels required for each class. We modify the base dataset by adding noise to all features  except the class feature. In the table~\ref{tab:synth}, you can see the structure of the manifold, the number of labeled samples in the target dataset, the noise added to the base dataset, as well as the mean accuracy with and without transfer learning. Based on the difference in mean accuracy with and without transfer learning, it can be seen how beneficial transfer learning can be for few-learning with manifold distributed data. Furthermore, the table illustrates how the manifold difference between the source and target datasets (noise) impacts the efficacy of transfer learning. 

\begin{table}[htb]
\centering
\setlength{\abovecaptionskip}{12pt}
\caption{\% Mean accuracy ($\mu\alpha$) with and without transfer learning (TL) on different manifold structures (1000 data points, 20 iterations) - $\sigma$ represents noise added. We also compare our approach with neural networks - domain adoption}
\label{tab:synth}
\small
\begin{tabular}{@{}>{\centering}p{1.5cm}>{\centering}p{1.5cm}>{\centering}p{1.2cm}>{\centering}p{1.2cm}>{\raggedleft}p{1cm}>{\raggedleft}p{1.7cm}>{\raggedleft}p{1.7cm}@{}}

\toprule
Shape             & Manifold & Sample(L) & $\sigma$- Noise  & $\mu\alpha \%$ & $\mu\alpha $ (TL-Our App)$\%$ & $\mu\alpha$ NN-DA$\%$ \\ 
\midrule
\multirow{4}{*}{\includegraphics[width=1.5cm, height=1.1cm]{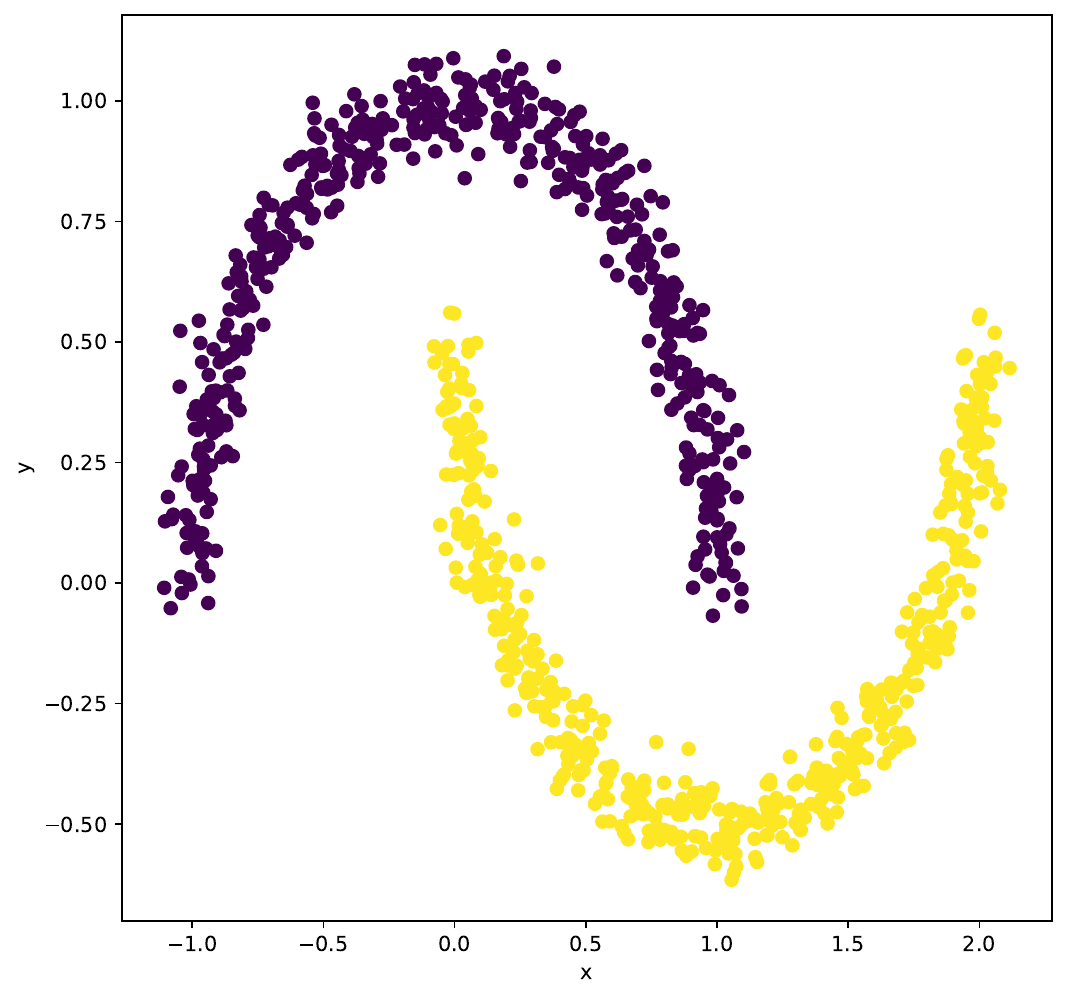}} & Moon       & 10        &     1     &       2.00      &  86.1                & 79.9               \\
                   & Moon       & 20        &   2       &  7.00           & 73.2                 & 68.3               \\
                   & Moon       & 30        &     3     &       9.10      &  60.9                & 60.0               \\
                   & Moon       & 40        &     4     &        22.2     & 56.7                 & 63.5              \\
\multirow{4}{*}{\includegraphics[width=1.5cm, height=1.1cm]{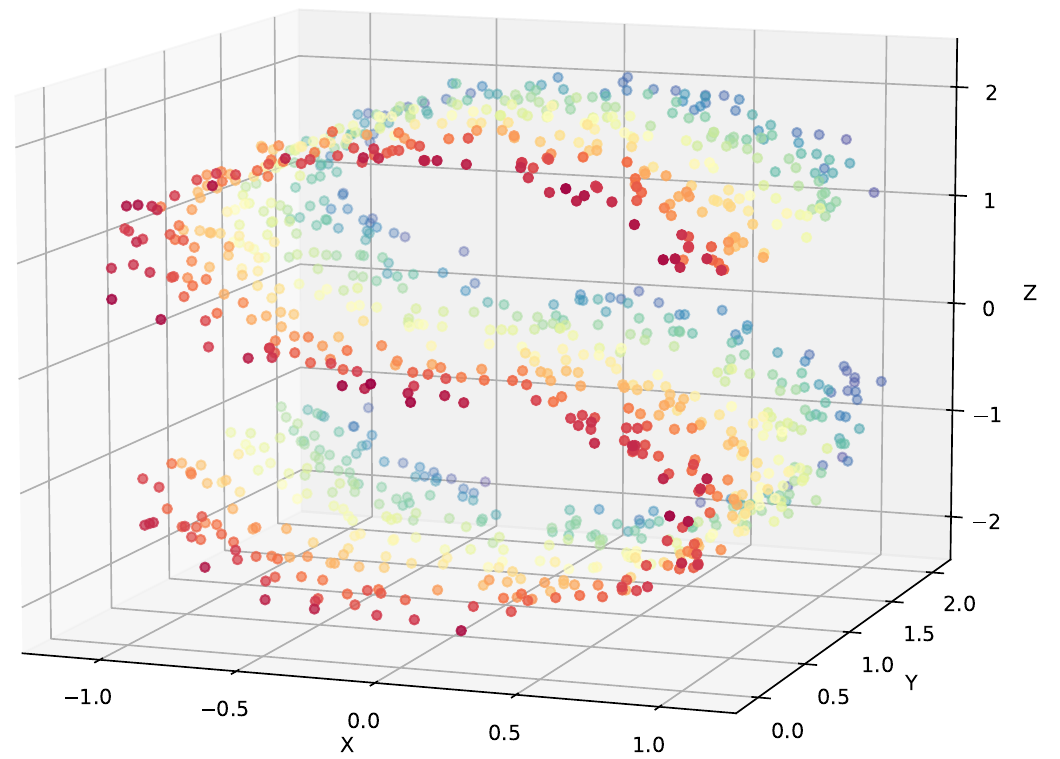}} & S Curve    & 10        &    1      &      14.1       &  39.4                & 33.9               \\
                   & S Curve    & 20        & 2         &  19.5           & 23.2                 & 25.7               \\
                   & S Curve    & 30        &    3      &         28.8    & 18.5                 & 16.9               \\
                   & S Curve    & 40        &       4   &     38.2        &  15.5                & 18.3               \\
\multirow{4}{*}{\includegraphics[width=1.5cm, height=1.1cm]{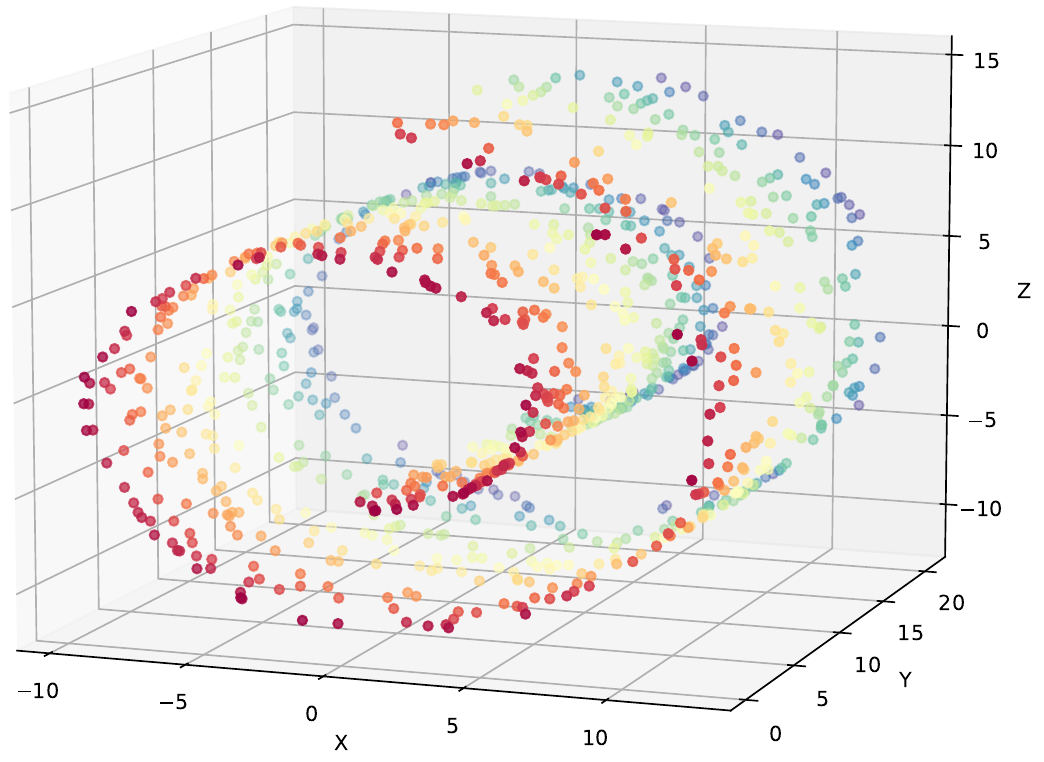}} & Swiss roll & 10        &  1        &      2.24       & 88.1                 & 22.5              \\
                   & Swiss roll & 20        &  2        &  7.00           & 74.8                 & 15.3              \\
                   & Swiss roll & 30        &     3     &     17.8        & 62.4                 & 12.2              \\
                   & Swiss roll & 40        &     4     &     32.2        &   58.2               & 9.8               \\
\bottomrule 
\end{tabular}

\end{table}

\subsection{Real world datasets}
We conducted experiments with banknotes, Pendigits, and Satlog datasets. You can access the datasets at (http://archive.ics.uci.edu). We generated two identical datasets: one as a target and the other as a source. The target dataset is modified by removing all labels except the number of labels required for each class. We modify the base dataset by adding noise to all features  except the class feature. In the table~\ref{tab:realworld}, you can see the structure of the manifold, the number of labeled samples in the target dataset, the noise added to the base dataset, as well as the mean accuracy with and without transfer learning. Based on the difference in mean accuracy with and without transfer learning, it can be seen how beneficial transfer learning can be for few-learning with manifold distributed data. Furthermore, the table illustrates how the manifold difference between the source and target datasets (noise) impacts the efficacy of transfer learning. 

\begin{table}[htb]
\centering
\setlength{\abovecaptionskip}{12pt}
\caption{\% Mean accuracy ($\mu\alpha$) with and without transfer learning (TL) on different manifold structures (1000 data points, 20 iterations) - $\sigma$ represents noise added. We also compare our approach with neural networks - domain adoption}
\label{tab:realworld}
\small
\begin{tabular}{@{}>{\centering}p{1.5cm}>{\centering}p{1.5cm}>{\centering}p{1.2cm}>{\centering}p{1.2cm}>{\raggedleft}p{1cm}>{\raggedleft}p{1.7cm}>{\raggedleft}p{1.7cm}@{}}
\toprule
Shape             & Dataset & Sample(L) & $\sigma$- Noise & $\mu\alpha \%$ & $\mu\alpha $ (TL - Our App)$\%$ & $\mu\alpha$ NN-DA$\%$ \\ 
\midrule
\multirow{4}{*}{\includegraphics[width=1.5cm, height=1.1cm]{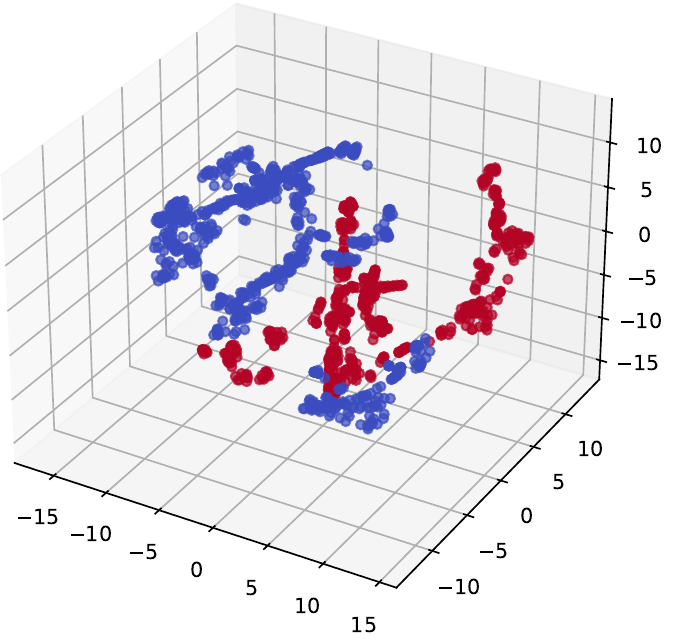}} & Banknotes       & 10        &     1     &       86.7      &  97.1                & 87.3               \\
                   & Banknotes       & 20        &   2       &  87.1           & 94.0                 & 88.5               \\
                   & Banknotes       & 30        &     3     &       87.7     &  90.8                & 86.8               \\
                   & Banknotes       & 40        &     4     &        88.8     & 87.2                 & 82.4              \\
\multirow{4}{*}{\includegraphics[width=1.5cm, height=1.1cm]{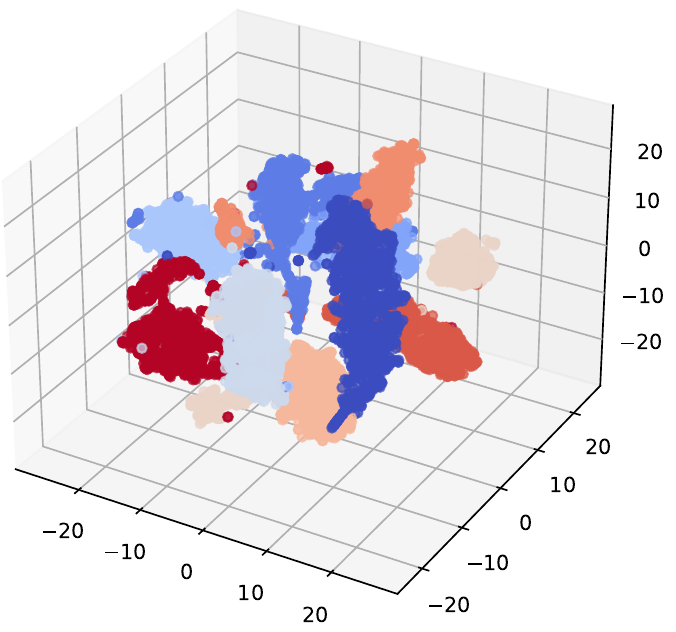}} & Pendigits& 10        &    1      &      87.4       &  99.1             & 2.4               \\
                   & Pendigits    & 20        & 2         &  90.9           & 99.2                 & 5.0               \\
                   & Pendigits    & 30        &    3      &         91.8    & 99.1                 & 9.7               \\
                   & Pendigits    & 40        &       4   &     93.8        &  98.7                & 17.3               \\
\multirow{4}{*}{\includegraphics[width=1.5cm, height=1.1cm]{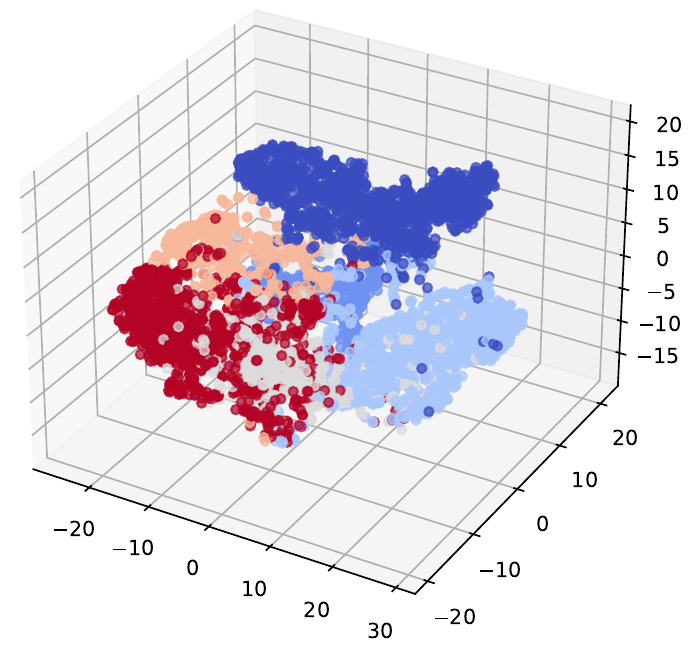}} & Satlog & 10        &  1        &      55.6       & 90.0                & 2.9               \\
                   & Satlog & 20        &  2        &  59.8           & 88.7                 & 10.7               \\
                   & Satlog & 30        &     3     &     67.6        & 88.5                 & 18.0               \\
                   & Satlog & 40        &     4     &     70.1        &   85.2               & 19.6               \\
\bottomrule 
\end{tabular}
\end{table}

\section{Pre-processing of image datasets with the superpixel centroid technique}
A similarity score is calculated after performing a significant amount of calculations on images that are similar. In scenarios where we datasets contain a large number of images to be compared, it is reasonable to perform a preprocessing stage by only comparing superpixel versions of these images~\cite{monti2017geometric}. Pre-processing enables the reduction of matching images to only relevant ones that can then be compared fully. This can be accomplished by utilizing a method that we propose and call Superpixel centriods. This method is designed to produce a centriod based superpixel image. A superpixel centroid image can be created by converting the image to a superpixel equivalent. Then, color each segment in the image by taking the colour of its centroid. For more information about how to create a superpixel centroid image, please refer to Algorithm~\ref{alg:superpixel}.  

Superpixel centroid images can be used for pre-processing purposes, but they can also be used for similarity calculations especially when high accuracy of similarity is not required. When such scenarios occur, it is essential to select the right number of segments for the superpixel centroid image. A smaller number of segments in the superpixel centroid image and a very high number of segments will decrease similarity accuracy. This is due to the fact that fewer numbers of segments cannot capture the manifold structure within an image effectively. Higher numbers can distort the manifold structure within the superpixel centroid equivalent image. Table xxx presents a similarity score based on our approach for an image and its superpixel equivalents of different segments. The table~\ref{tab:superpixel} shows clearly how the number of segments affects the similarity measure. 

\begin{table}
    \centering
    \setlength{\abovecaptionskip}{12pt}
    \caption{Impact of different Superpixel Centroid Image Sizes on Similarity Measures Between Two Bird Images}
    \label{tab:superpixel}
    \begin{tabular}{>{\centering\arraybackslash}m{1.5cm} >{\centering\arraybackslash}m{1.5cm} >{\centering\arraybackslash}m{1.5cm} >{\centering\arraybackslash}m{1.5cm}}
        \toprule
        \textbf{A} & \textbf{B} & \textbf{Sup-C Size} & \textbf{Similarity} \\
        \midrule
        \includegraphics[width=1.5cm, height=1.5cm]{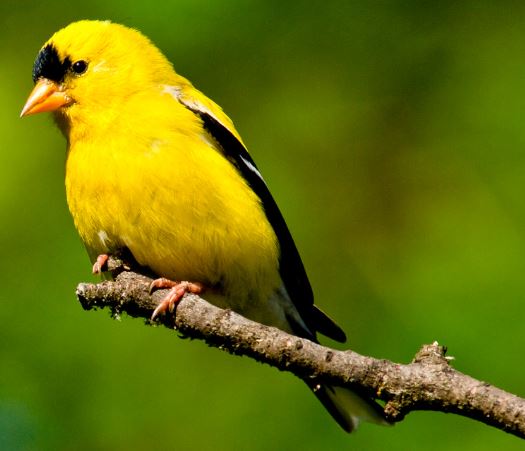} & \includegraphics[width=1.5cm, height=1.5cm]{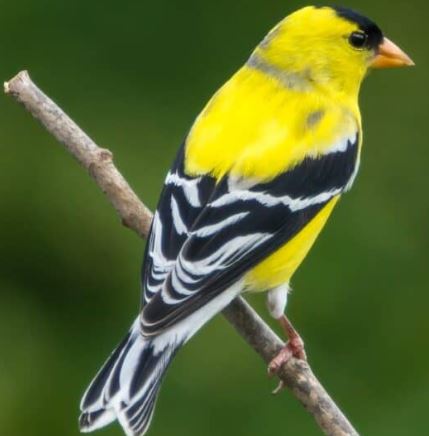} & Original Size & 45.40 \\
        \includegraphics[width=1.5cm, height=1.5cm]{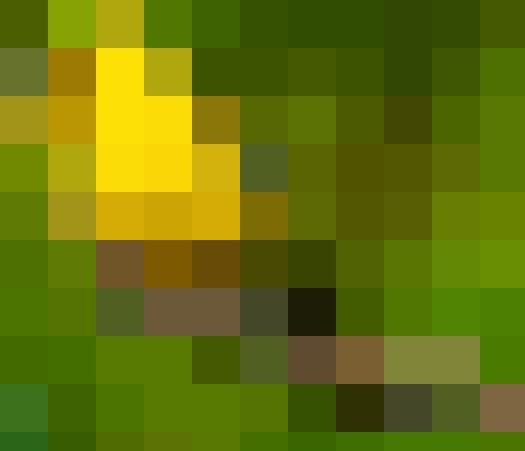} & \includegraphics[width=1.5cm, height=1.5cm]{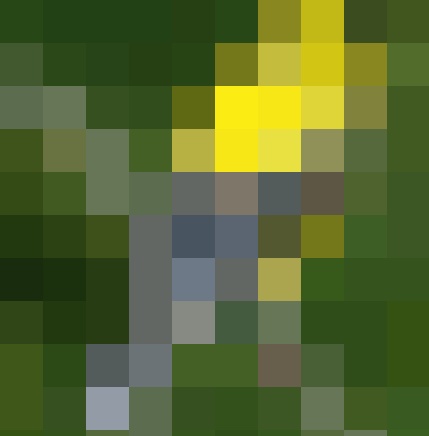} & 10 x 10 & 1176.42 \\
        \includegraphics[width=1.5cm, height=1.5cm]{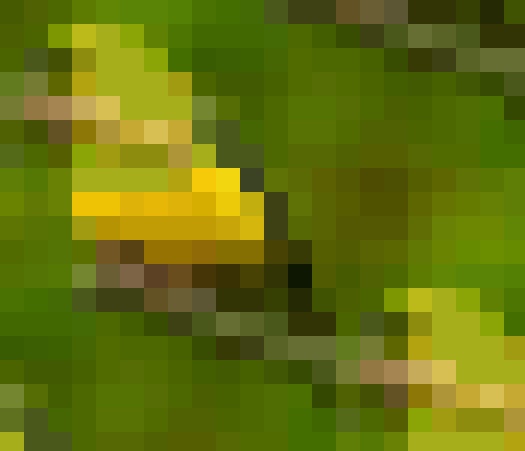} & \includegraphics[width=1.5cm, height=1.5cm]{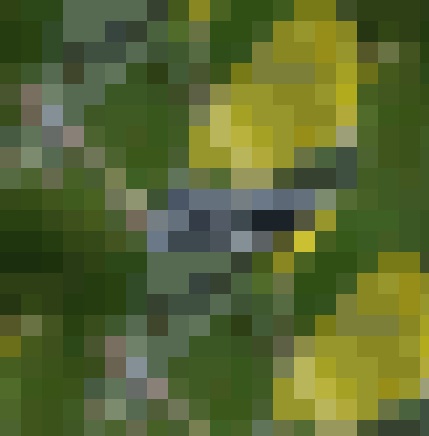} & 20 x 20 & 94.91 \\
        \includegraphics[width=1.5cm, height=1.5cm]{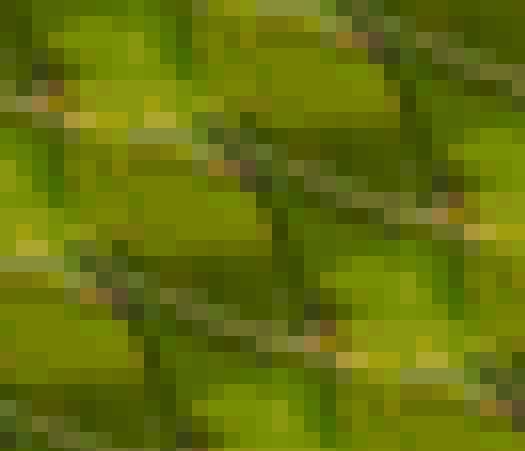} & \includegraphics[width=1.5cm, height=1.5cm]{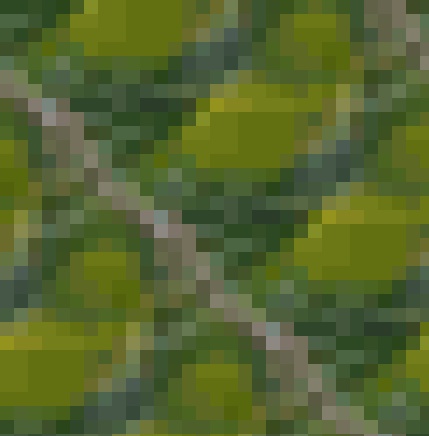} & 30 x 30 & 387.07 \\
        \bottomrule
    \end{tabular}
\end{table}

\begin{algorithm}
\caption{Convert Image to Superpixel Centroid Image}
\label{alg:superpixel}
\begin{algorithmic}[1]
\item \textbf{Input:} Image $I$, number of superpixel segments $N$
\item \textbf{Output:} Superpixel Image $I$

\item \textbf{Step 1: Load and Preprocess Image}
\begin{enumerate}
    \item Load the input image $I$
    \item Convert the image $I$ to LAB color space to get $lab\_image$
\end{enumerate}

\item \textbf{Step 2: Initialize and Apply Superpixel Segmentation}
\begin{enumerate}
    \item Initialize the superpixel segmentation function
    \item \textbf{Function} \texttt{superpixels(image,$N$)}
    \begin{enumerate}
        \item Initialize superpixel image with zeros
        \item Compute step size as $step = \sqrt{(image\_height \times image\_width) / N}$
        \item Assign labels to superpixel by starting with a $label = 0$
        \item \textbf{for} $y$ from $0$ to $image\_height$ with step $step$
        \begin{enumerate}
            \item \textbf{for} $x$ from $0$ to $image\_width$ with step $step$
            \begin{enumerate}
                \item Set $segments\_slic[y:y + step, x:x + step] = label$
                \item Increment $label$
            \end{enumerate}
            \item \textbf{end for}
        \end{enumerate}
        \item \textbf{end for}
        \item \textbf{return} $segments\_slic$
    \end{enumerate}
    \item \textbf{End Function}
    \item Call $segments\_slic = superpixels(lab\_image, N)$
\end{enumerate}

\item \textbf{Step 3: Apply K-Means Clustering}
\begin{enumerate}
    \item Apply K-Means clustering on reshaped $lab\_image$ with $N$ clusters to get $palette\_colors\_lab$
\end{enumerate}

\item \textbf{Step 4: Create Superpixel Centroid Image}
\begin{enumerate}
    \item Initialize $superpixel\_centroids$ with zeros of the same size as $I$
    \item \textbf{for each} unique segment label $segment$ in $segments\_slic$
    \begin{enumerate}
        \item Create $mask$ for pixels belonging to the current $segment$
        \item Compute the centroid color $centroid\_color\_lab$ of the superpixel in LAB space
        \item Find the closest color in $palette\_colors\_lab$ to $centroid\_color\_lab$
        \item Set the color of all pixels in $superpixel\_centroids$ belonging to $segment$ to the closest color
    \end{enumerate}
    \item \textbf{end for}
    \item Convert $superpixel\_centroids$ from LAB to BGR color space
\end{enumerate}

\item \textbf{Step 5: Return Results}
\begin{enumerate}
    \item \textbf{return} $superpixel\_centroids$
\end{enumerate}

\end{algorithmic}
\end{algorithm}


\section{Reinforcement Learning: Broadening Capabilities and Applications}
\label{RL}
Manifold recognition of images can be extremely useful in reinforcement learning, especially in tasks where visual input plays a critical role, like robotics, autonomous driving, and computer vision games. In reinforcement learning, manifold recognition refers to the underlying structures or patterns within a high-dimensional image space. The following are some of the ways in which manifold recognition may assist reinforcement learning:  

\begin{enumerate}

\item Using manifold learning techniques, meaningful features can be extracted from high-dimensional image data while maintaining essential information about the image's structure and characteristics~\cite{zhu2018image}. Learning and generalisation can then be improved by using these extracted features as inputs to reinforcement learning algorithms.  

\item By recognising the manifold structure within images, it is possible to produce state representations that are more compact and informative in reinforcement learning tasks. By reducing state space complexity and focusing learning efforts on relevant features and patterns, reinforcement learning algorithms can increase their efficiency~\cite{li2017manifold}.

\item Manifold recognition can help understand semantic relationships and hierarchies within image data~\cite{he2004learning}. It can enable reinforcement learning agents to learn meaningful abstractions and representations of the environment, which would lead to better decision-making and the generation of appropriate behaviours.  

\item In image-based reinforcement learning tasks, manifold recognition can provide guidance for exploration strategies~\cite{srinivasan2015improving}. By focusing exploration efforts in areas of image space where uncertainty or novel information are likely to be found, agents will be able to achieve greater efficiency in learning and improve generalisation to unknown situations.

\item A manifold-based representation can enhance the robustness of reinforcement learning algorithms to noise, variation, and distortions in image data~\cite{wang2010improved}. It is possible for agents to learn robust and stable policies that generalise well across different visual conditions by capturing the underlying structure and ignoring irrelevant variations.  
\end{enumerate}

\begin{figure}[htb]
    \centering
     \caption{MiniImageNet Dataset: A reinforcement agent trained on the manifold structure of one image can identify and classify all other images.}
    \label{tab:reinforcement}
    \begin{tabular}{cc}


    \begin{tikzpicture}

        \node[anchor=center] (center1) at (0, 0) {\includegraphics[width=1cm, height=1cm]{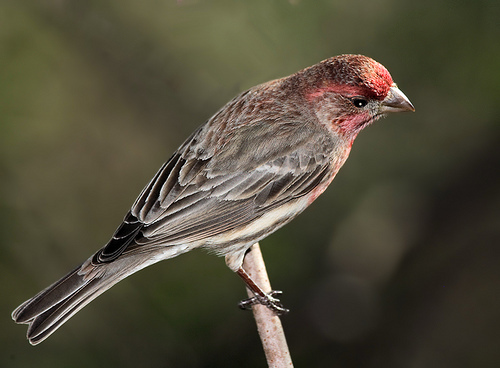}};
        
        \node[anchor=center] at (36:2) {\includegraphics[width=1cm, height=1cm]{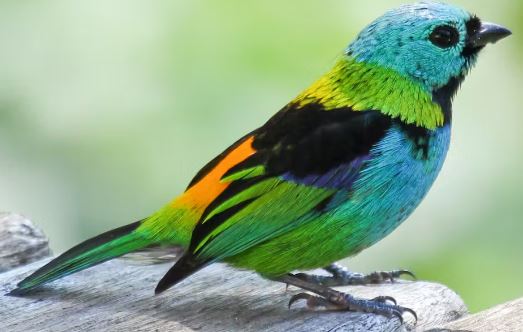}};
        \node[anchor=center] at (72:2) {\includegraphics[width=1cm, height=1cm]{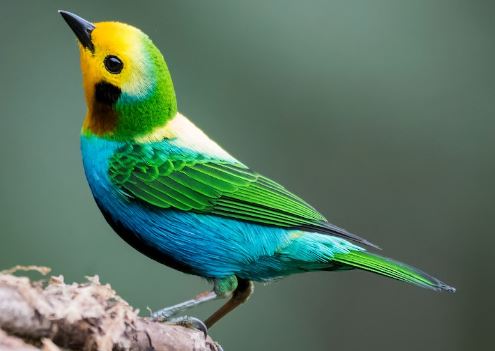}};
        \node[anchor=center] at (108:2) {\includegraphics[width=1cm, height=1cm]{Birds/Bird_yellow_1.JPG}};
        \node[anchor=center] at (144:2) {\includegraphics[width=1cm, height=1cm]{Birds/Bird_yellow_2.JPG}};
        \node[anchor=center] at (180:2) {\includegraphics[width=1cm, height=1cm]{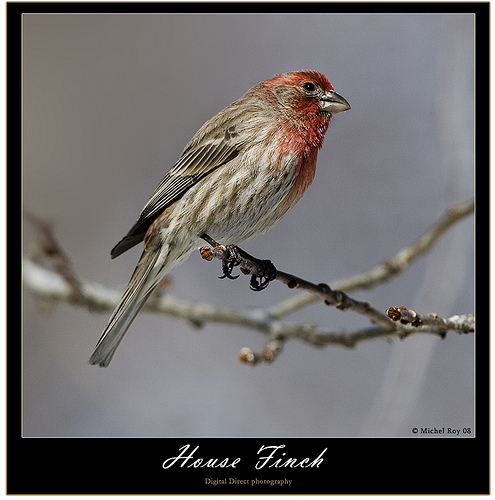}};
        \node[anchor=center] at (216:2) {\includegraphics[width=1cm, height=1cm]{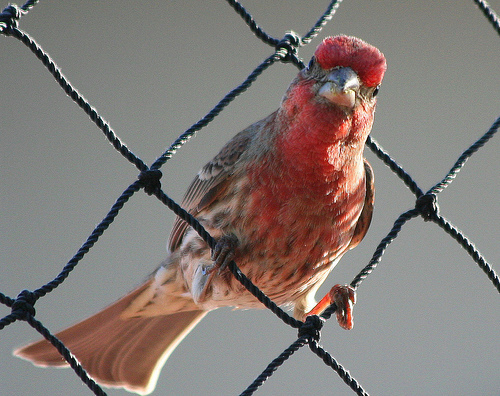}};
        \node[anchor=center] at (252:2) {\includegraphics[width=1cm, height=1cm]{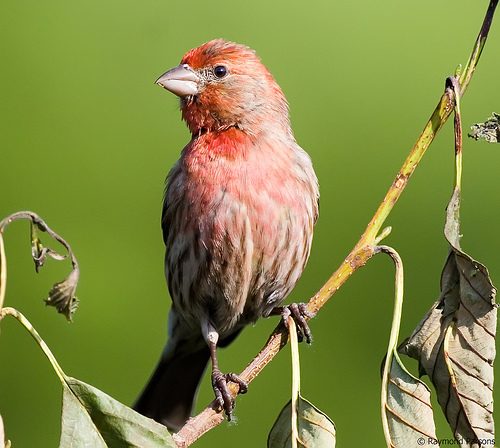}};
        \node[anchor=center] at (288:2) {\includegraphics[width=1cm, height=1cm]{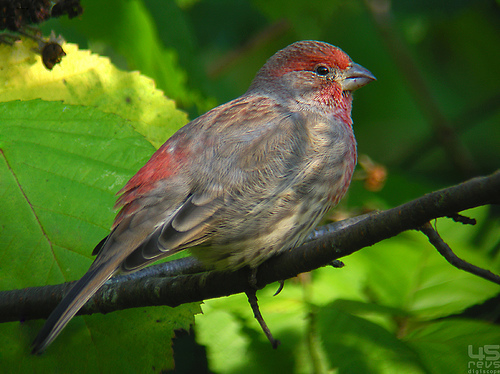}};
        \node[anchor=center] at (324:2) {\includegraphics[width=1cm, height=1cm]{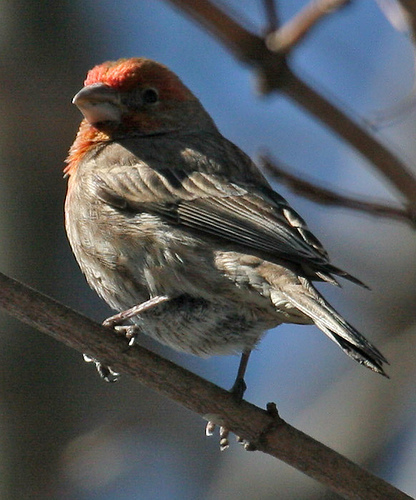}};
        \node[anchor=center] at (360:2) {\includegraphics[width=1cm, height=1cm]{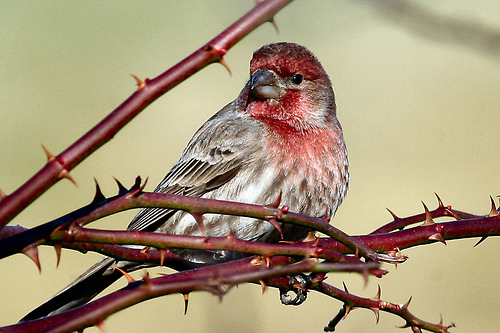}};

        \foreach \i in {36, 72, 108, 144, 180, 216, 252, 288, 324, 360} {
            \draw[->, thick] (center1) -- (\i:1);
        }
    \end{tikzpicture}

&
    \begin{tikzpicture}
         \node[anchor=center] (center1) at (0, 0) {\includegraphics[width=1cm, height=1cm]{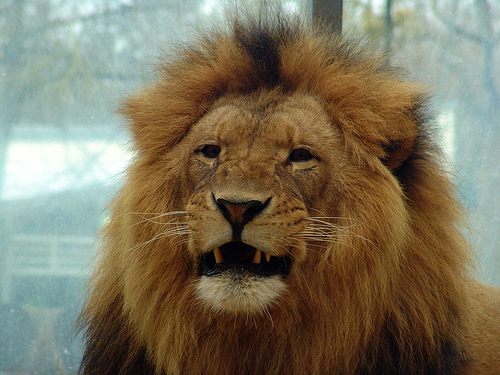}};
        
        \node[anchor=center] at (36:2) {\includegraphics[width=1cm, height=1cm]{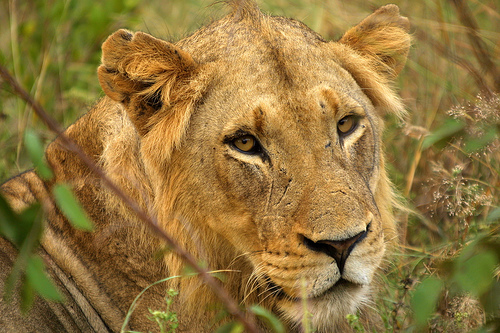}};
        \node[anchor=center] at (72:2) {\includegraphics[width=1cm, height=1cm]{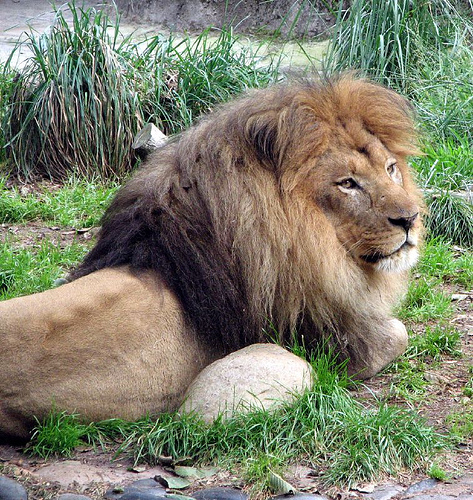}};
        \node[anchor=center] at (108:2) {\includegraphics[width=1cm, height=1cm]{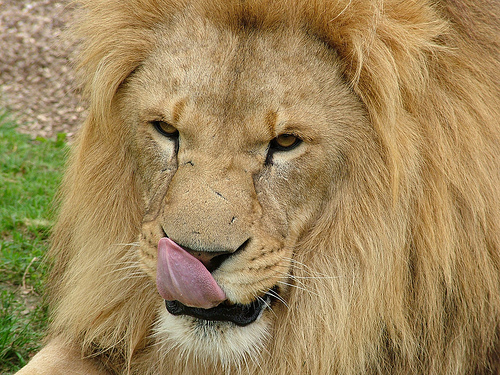}};
        \node[anchor=center] at (144:2) {\includegraphics[width=1cm, height=1cm]{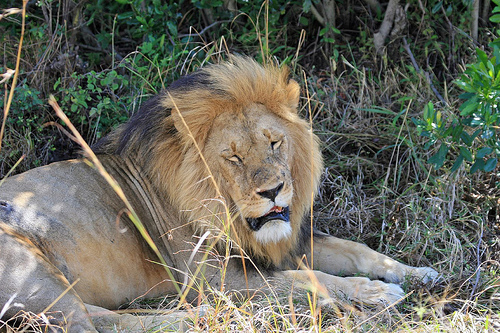}};
        \node[anchor=center] at (180:2) {\includegraphics[width=1cm, height=1cm]{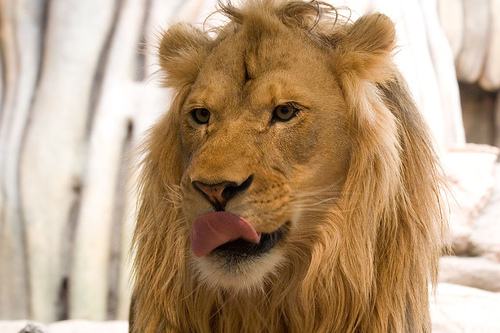}};
        \node[anchor=center] at (216:2) {\includegraphics[width=1cm, height=1cm]{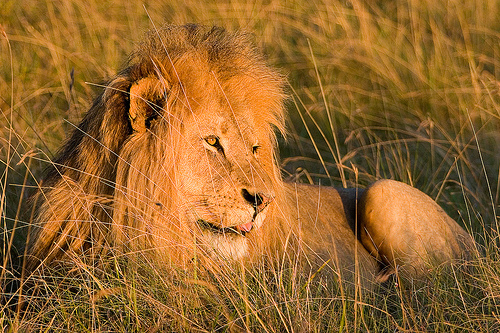}};
        \node[anchor=center] at (252:2) {\includegraphics[width=1cm, height=1cm]{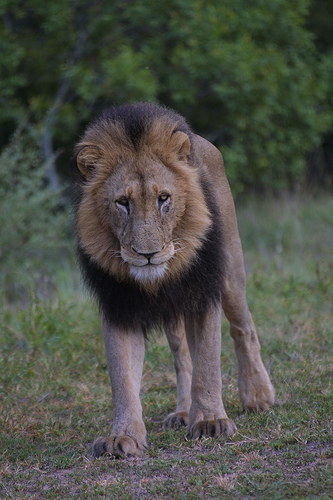}};
        \node[anchor=center] at (288:2) {\includegraphics[width=1cm, height=1cm]{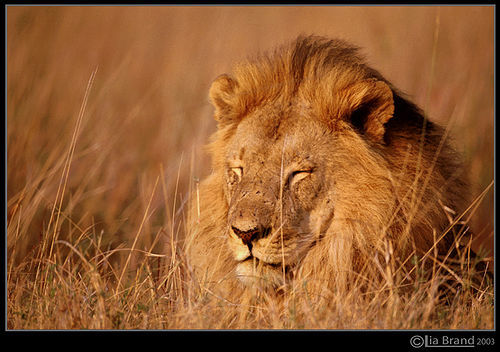}};
        \node[anchor=center] at (324:2) {\includegraphics[width=1cm, height=1cm]{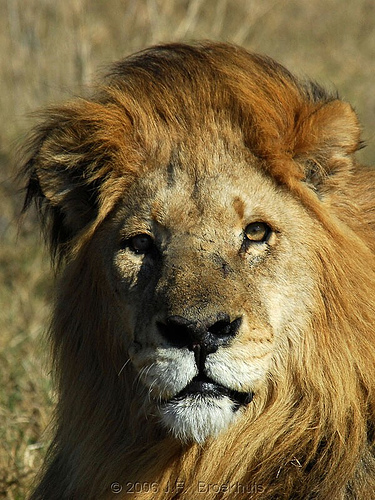}};
        \node[anchor=center] at (360:2) {\includegraphics[width=1cm, height=1cm]{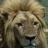}};

        \foreach \i in {36, 72, 108, 144, 180, 216, 252, 288, 324, 360} {
            \draw[->, thick] (center1) -- (\i:1);
        }
    \end{tikzpicture}


\end{tabular}
\end{figure}

Using our approach to measure similarity in the above context can enhance image-intensive reinforcement learning through measuring the similarity between an image and a database of known images. Table~\ref{tab:reinforcement} illustrates how one picture depicting a bird and a lion can be used to identify all other similar images of the same class. This is based on its similarity to that picture's manifold structure. Due to the fact that our method compares only manifold structures, not the entire image, we have minimised noise, reduced exploration areas, reduced the complexity of comparing the entire image, focused on relevant features and patterns, and achieved a more general approach to similarity measurement.     

Overall, a manifold recognition approach to reinforcement learning is a valuable method in various image-intensive reinforcement learning applications, allowing for enhanced learning efficiency, better generalisation, and higher robustness to noise.

\section{Conclusions}
The development of a classifier with high mean accuracy from a manifold distributed dataset with limited labels is not an easy task. In order for transfer learning to be effective, the manifold structure of the base dataset must be similar to the target dataset. As a means of solving these problems, this paper proposes a novel method for calculating similarity between two manifold distributed datasets. It also proposes a novel transfer learning algorithm in few-shot learning scenarios with manifold distributed datasets. Graphs and random walks are used in our method for calculating similarity between manifold structures. Using the manifold information along with the label distribution from the base manifold distributed dataset and the limited labels from the target dataset, we can construct a high mean accuracy classifier. 

We also discuss a superpixel centroid-based approach to pre-processing image datasets in this paper in order to achieve processing efficiency in large image datasets. Finally, we discuss the application of the approach to reinforcement learning as well as future directions for this study.
\\\\

\bibliography{references.bib}

\begin{thebibliography}{33}
\providecommand{\natexlab}[1]{#1}
\providecommand{\url}[1]{{#1}}
\providecommand{\urlprefix}{URL }
\providecommand{\doi}[1]{\url{https://doi.org/#1}}
\providecommand{\eprint}[2][]{\url{#2}}
 \bibcommenthead

\bibitem[{Cai et~al(2010)Cai, He, Han, and Huang}]{cai2010graph}
Cai D, He X, Han J, et~al (2010) Graph regularized nonnegative matrix factorization for data representation. IEEE transactions on pattern analysis and machine intelligence 33(8):1548--1560

\bibitem[{Cao et~al(2010)Cao, Pan, Zhang, Yeung, and Yang}]{cao2010adaptive}
Cao B, Pan SJ, Zhang Y, et~al (2010) Adaptive transfer learning. In: proceedings of the AAAI Conference on Artificial Intelligence, pp 407--412

\bibitem[{Dvornik et~al(2019)Dvornik, Schmid, and Mairal}]{dvornik2019diversity}
Dvornik N, Schmid C, Mairal J (2019) Diversity with cooperation: Ensemble methods for few-shot classification. In: Proceedings of the IEEE/CVF international conference on computer vision, pp 3723--3731

\bibitem[{Elsken et~al(2020)Elsken, Staffler, Metzen, and Hutter}]{elsken2020meta}
Elsken T, Staffler B, Metzen JH, et~al (2020) Meta-learning of neural architectures for few-shot learning. In: Proceedings of the IEEE/CVF conference on computer vision and pattern recognition, pp 12365--12375

\bibitem[{Finn et~al(2017)Finn, Abbeel, and Levine}]{finn2017model}
Finn C, Abbeel P, Levine S (2017) Model-agnostic meta-learning for fast adaptation of deep networks. In: International conference on machine learning, PMLR, pp 1126--1135

\bibitem[{Ganin et~al(2016)Ganin, Ustinova, Ajakan, Germain, Larochelle, Laviolette, Marchand, and Lempitsky}]{ganin2016domain}
Ganin Y, Ustinova E, Ajakan H, et~al (2016) Domain-adversarial training of neural networks. The journal of machine learning research 17(1):2096--2030

\bibitem[{Goodfellow et~al(2020)Goodfellow, Pouget-Abadie, Mirza, Xu, Warde-Farley, Ozair, Courville, and Bengio}]{goodfellow2020generative}
Goodfellow I, Pouget-Abadie J, Mirza M, et~al (2020) Generative adversarial networks. Communications of the ACM 63(11):139--144

\bibitem[{He et~al(2016)He, Zhang, Ren, and Sun}]{he2016deep}
He K, Zhang X, Ren S, et~al (2016) Deep residual learning for image recognition. In: Proceedings of the IEEE conference on computer vision and pattern recognition, pp 770--778

\bibitem[{He et~al(2004)He, Ma, and Zhang}]{he2004learning}
He X, Ma WY, Zhang HJ (2004) Learning an image manifold for retrieval. In: Proceedings of the 12th annual ACM international conference on Multimedia, pp 17--23

\bibitem[{Jiang et~al(2020)Jiang, Huang, Geng, and Deng}]{jiang2020multi}
Jiang W, Huang K, Geng J, et~al (2020) Multi-scale metric learning for few-shot learning. IEEE Transactions on Circuits and Systems for Video Technology 31(3):1091--1102

\bibitem[{Krizhevsky et~al(2012)Krizhevsky, Sutskever, and Hinton}]{krizhevsky2012imagenet}
Krizhevsky A, Sutskever I, Hinton GE (2012) Imagenet classification with deep convolutional neural networks. Advances in neural information processing systems 25

\bibitem[{Li et~al(2017)Li, Liu, and Wang}]{li2017manifold}
Li H, Liu D, Wang D (2017) Manifold regularized reinforcement learning. IEEE transactions on neural networks and learning systems 29(4):932--943

\bibitem[{Monti et~al(2017)Monti, Boscaini, Masci, Rodola, Svoboda, and Bronstein}]{monti2017geometric}
Monti F, Boscaini D, Masci J, et~al (2017) Geometric deep learning on graphs and manifolds using mixture model cnns. In: Proceedings of the IEEE conference on computer vision and pattern recognition, pp 5115--5124

\bibitem[{Qayyumi et~al(2023)Qayyumi, Park, and Obst}]{qayyumi2023few}
Qayyumi SW, Park LA, Obst O (2023) Few-shot and transfer learning with manifold distributed datasets. In: Australasian Conference on Data Science and Machine Learning, Springer, pp 137--149

\bibitem[{Reddy et~al(2018)Reddy, Viswanath, and Reddy}]{reddy2018semi}
Reddy Y, Viswanath P, Reddy BE (2018) Semi-supervised learning: A brief review. Int J Eng Technol 7(1.8):81

\bibitem[{Ren et~al(2018)Ren, Triantafillou, Ravi, Snell, Swersky, Tenenbaum, Larochelle, and Zemel}]{ren2018meta}
Ren M, Triantafillou E, Ravi S, et~al (2018) Meta-learning for semi-supervised few-shot classification. arXiv preprint arXiv:180300676

\bibitem[{Robb et~al(2020)Robb, Chu, Kumar, and Huang}]{robb2020few}
Robb E, Chu WS, Kumar A, et~al (2020) Few-shot adaptation of generative adversarial networks. arXiv preprint arXiv:201011943

\bibitem[{Rostami et~al(2019)Rostami, Kolouri, Eaton, and Kim}]{rostami2019deep}
Rostami M, Kolouri S, Eaton E, et~al (2019) Deep transfer learning for few-shot sar image classification. Remote Sensing 11(11):1374

\bibitem[{Simonyan and Zisserman(2014)}]{simonyan2014very}
Simonyan K, Zisserman A (2014) Very deep convolutional networks for large-scale image recognition. arXiv preprint arXiv:14091556

\bibitem[{Snell et~al(2017)Snell, Swersky, and Zemel}]{snell2017prototypical}
Snell J, Swersky K, Zemel R (2017) Prototypical networks for few-shot learning. In: Advances in Neural Information Processing Systems, pp 4077--4087

\bibitem[{Srinivasan et~al(2015)Srinivasan, Talvitie, and Bowling}]{srinivasan2015improving}
Srinivasan S, Talvitie E, Bowling M (2015) Improving exploration in uct using local manifolds. In: Proceedings of the AAAI Conference on Artificial Intelligence

\bibitem[{Tu et~al(2014)Tu, Cao, Yang, and Kasabov}]{tu2014novel}
Tu E, Cao L, Yang J, et~al (2014) A novel graph-based k-means for nonlinear manifold clustering and representative selection. Neurocomputing 143:109--122

\bibitem[{Tu et~al(2016)Tu, Zhang, Zhu, Yang, and Kasabov}]{tu2016graph}
Tu E, Zhang Y, Zhu L, et~al (2016) A graph-based semi-supervised k nearest-neighbor method for nonlinear manifold distributed data classification. Information Sciences 367--368:673--688

\bibitem[{Vinyals et~al(2016)Vinyals, Blundell, Lillicrap, Wierstra et~al}]{vinyals2016matching}
Vinyals O, Blundell C, Lillicrap T, et~al (2016) Matching networks for one shot learning. In: Advances in Neural Information Processing Systems, pp 3630--3638

\bibitem[{Vishwanathan et~al(2010)Vishwanathan, Schraudolph, Kondor, and Borgwardt}]{vishwanathan2010graph}
Vishwanathan SVN, Schraudolph NN, Kondor R, et~al (2010) Graph kernels. Journal of Machine Learning Research 11:1201--1242

\bibitem[{Wang and Huang(2010)}]{wang2010improved}
Wang GB, Huang LP (2010) An improved noise reduction algorithm based on manifold learning and its application to signal noise reduction. Applied Mechanics and Materials 26:653--656

\bibitem[{Wang et~al(2020)Wang, Yao, Kwok, and Ni}]{wang2020generalizing}
Wang Y, Yao Q, Kwok JT, et~al (2020) Generalizing from a few examples: A survey on few-shot learning. ACM computing surveys (csur) 53(3):1--34

\bibitem[{Weiss et~al(2016)Weiss, Khoshgoftaar, and Wang}]{weiss2016survey}
Weiss K, Khoshgoftaar TM, Wang D (2016) A survey of transfer learning. Journal of Big data 3(1):1--40

\bibitem[{Zhang et~al(2014)Zhang, Wen, Wang, and Jiang}]{zhang2014semi}
Zhang Y, Wen J, Wang X, et~al (2014) Semi-supervised learning combining co-training with active learning. Expert Systems with Applications 41(5):2372--2378

\bibitem[{Zhou et~al(2021)Zhou, Zheng, Tang, Li, and Yang}]{zhou2021flipda}
Zhou J, Zheng Y, Tang J, et~al (2021) Flipda: Effective and robust data augmentation for few-shot learning. arXiv preprint arXiv:210806332

\bibitem[{Zhu et~al(2018)Zhu, Liu, Cauley, Rosen, and Rosen}]{zhu2018image}
Zhu B, Liu JZ, Cauley SF, et~al (2018) Image reconstruction by domain-transform manifold learning. Nature 555(7697):487--492

\bibitem[{Zhu(2005)}]{zhu2005semi}
Zhu XJ (2005) Semi-supervised learning literature survey

\bibitem[{Zoph et~al(2020)Zoph, Ghiasi, Lin, Cui, Liu, Cubuk, and Le}]{zoph2020rethinking}
Zoph B, Ghiasi G, Lin TY, et~al (2020) Rethinking pre-training and self-training. Advances in neural information processing systems 33:3833--3845

\end{thebibliography}

\end{document}